# Incremental Tube Construction for Human Action Detection

Harkirat Singh Behl[1]
harkirat@robots.ox.ac.uk

Michael Sapienza[2]
m.sapienza@samsung.com

Gurkirt Singh[3]
gurkirt.singh-2015@brookes.ac.uk

Suman Saha[3]
suman.saha-2014@brookes.ac.uk

Fabio Cuzzolin[3]
fabio.cuzzolin@brookes.ac.uk

Philip H. S. Torr[1]
phst@robots.ox.ac.uk

[1] Department of Engineering Science
University of Oxford
Oxford, UK

[2] Think Tank Team
Samsung Research America
Mountain View, CA

[3] Dept. of Computing and
Communication Technologies
Oxford Brookes University
Oxford, UK

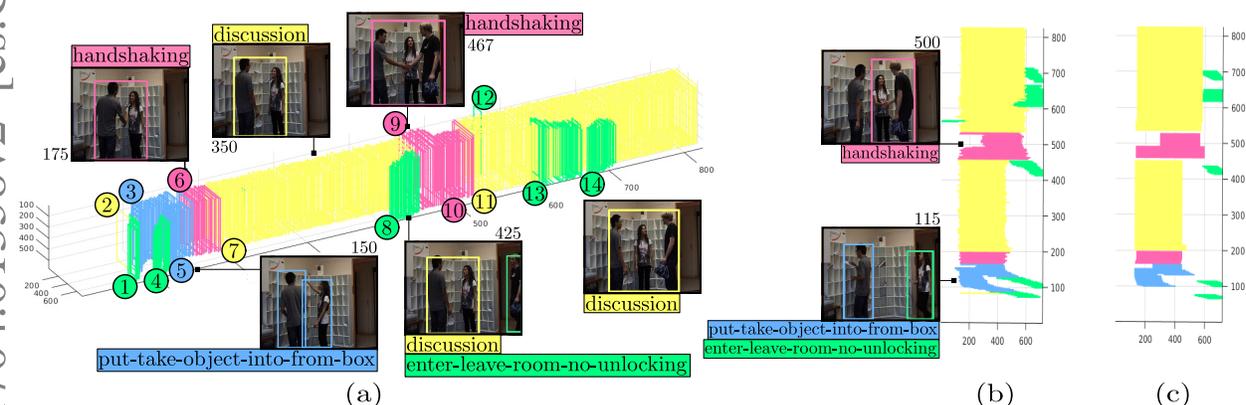

Figure 1: **(a)** *Illustrative results on a video sequence from the LIRIS-HARL dataset [23]. Two people enter a room and put/take an object from a box (frame 150). They then shake hands (frame 175) and start having a discussion (frame 350). In frame 450, another person enters the room, shakes hands, and then joins the discussion. Each action tube instance is numbered and coloured according to its action category. We selected this video to show that our tube construction algorithm can handle very complex situations in which multiple distinct action categories occur in sequence and at concurrent times.* **(b)** *Action tubes drawn as viewed from above, compared to* **(c)** *the ground truth action tubes.*

## Abstract

Current state-of-the-art action detection systems are tailored for offline batch-processing applications. However, for online applications like human-robot interaction, current systems fall short. In this work, we introduce a real-time and online joint-labelling and association algorithm for action detection that can incrementally construct space-time action tubes on the most challenging untrimmed action videos in which different action categories occur concurrently. In contrast to previous methods, we solve the linking, action labelling and temporal localization problems jointly in a single pass. We demonstrate superior online association accuracy and speed (1.8ms per frame) as compared to the current state-of-the-art offline and online systems.

# 1 Introduction

Detecting human actions has been defined as the task of automatically predicting the start, end and spatial extent of various actions [10, 21, 23] by predicting sets of connected windows in time (called tubes) in which each action is enclosed, as illustrated in Fig.1. Human action detection has gained huge popularity in the computer vision community due to its broad range of exciting applications. On the one hand, it is useful in 'offline' batch applications such as surveillance and the retrieval of video content in huge video collections. On the other hand, it can be used for online[1] human-robot interaction [19], an application in which both agents require instantaneous feedback and in which frame processing needs to be real-time and incremental.

Current state-of-the-art action detection methods have made remarkable progress on the aforementioned batch processing applications [14, 17, 22], by dividing the action detection task into two steps: i) the extraction of independent frame-level action detection-windows or action tubelets from a sequence of frames, and ii) the linking (association) of the detection-windows or tubelets to form action tubes. Due to the recent success of deep CNNs in object detection, action detection-windows and action tubelets have improved drastically [11, 14, 17, 20]. The linking, however, still has several shortcomings.

Firstly, the linking is done by treating the video as a 3D block of frames (offline) [7, 14, 17]. In this way, one can solve the problem globally. However, they cannot be used for online applications.

Secondly, previous methods divide the problem into independent parts because that makes it easier to solve. In particular, the association and temporal localisation are performed as two separate optimisation steps [14, 17, 18, 20]. This multi-part optimisation process is computationally redundant and expensive, which also makes it harder to scale to a large number of action classes. Moreover, the action tubes are constructed for each action class independently [11, 14, 17, 18, 20]. This can cause several action tubes with different action classes to overlap in the same space-time region (see Fig.3 and Fig.4), even though only one action might be happening. Furthermore, a single human may be represented by several tubes; and there is no way to tell whether a single human is performing multiple actions or multiple humans are performing individual actions. The effects of these shortcomings can be seen in Figs. 2b, 3 and 4.

**Contributions.**    In this work, we tackle real-time human action detection applications where multiple tubes with different action classes are present concurrently. We propose a novel linking algorithm called OJLA (Online Joint Labelling and Association), that is able to construct and update action tubes as each new frame is added. Our proposed algorithm does away with multiple optimisation passes for association, temporal localisation and labelling [11, 17, 18, 20]. Instead we formulate a novel cost function which solves all of these tasks jointly and incrementally in a single pass.

This implies that we do not perform action detection separately for each class. For scenarios where only one human action is taking place in a space-time location, which is the case in UCF-101, JHMDB-21 and LIRIS-HARL, our work outputs several human-centered (non-overlapping) action tubes, where each tube can take a single label. Thus avoiding the problem of detecting multiple co-located action tubes with different classes (see Fig. 3). For less constrained applications (single human performing multiple actions at the same time, for e.g walking and talking), our work can also output several human-centered (non-overlapping) action tubes, with each tube taking multiple labels.

---

[1] By online we specifically mean that the algorithm is capable of processing its input frame-by-frame, without having the entire input available from the start.

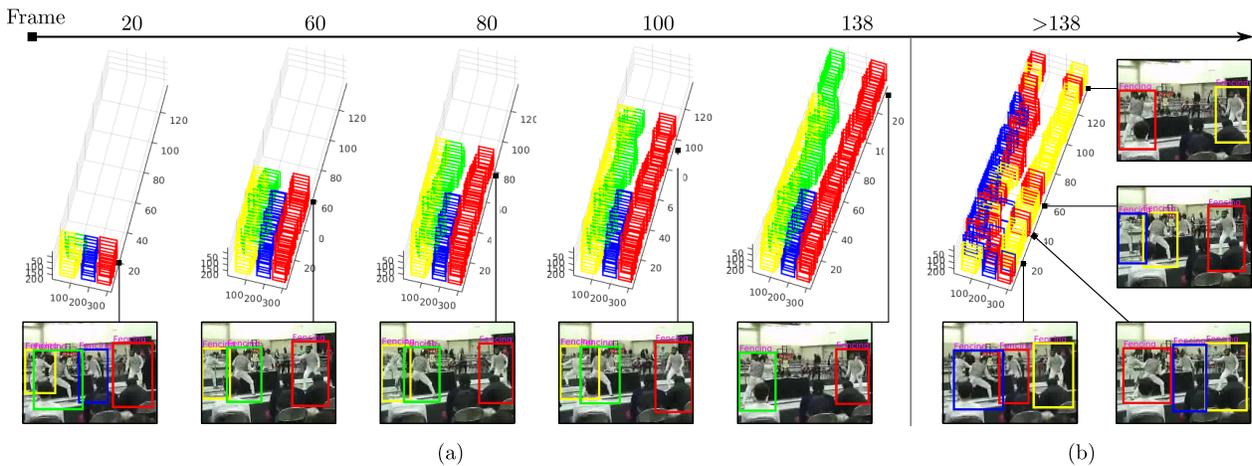

*Figure 2:* An illustration of the difference between **(a)** our online, incremental algorithm for constructing action tubes, and **(b)** one of the state-of-the-art offline methods by [17], using the exact same scores coming from a two-stream CNN. The video from UCF-101 contains four humans competing in a fencing tournament. Each colour indicates a different action tube present in the video. Notice that in **(a)**, at each new frame in time, our method is able to predict the detection-window association and class of each currently active action tube. Solving for the association and class assignment in one pass allows us to reduce the window switching effects **(b)** observed in the results by prior art [17].

We demonstrate that our action tube construction algorithm outperforms previous batch methods [14, 17] in terms of accuracy and speed (association takes ~1.8ms per frame or ~550 fps when averaged over all videos on UCF101), by virtue of our representation and the simplified single-pass optimisation coupled with an appropriate cost function.

Finally, we demonstrate a real-time action detection system (§ 5), which combines the real-time frame-level window detector of [13] and our OJLA algorithm, and outperforms current state-of-the-art offline methods. We have released the code[2] and a video[3] of this work.

## 2 Related Work

**Multiple action detection in space and time.** Several works attempt to localise actions in space alone using a single-frame detector [3, 25], or in time alone [4, 6]. Here we focus on space-time action detection, defined by a set of continuous sequences of 2D detection-windows in time to form action tubes. One of the first works on action detection in realistic scenarios was pioneered by Laptev *et al.* [12] in which a boosted 3D space-time window classifier combined with a single frame action detector. Note that Yu *et al.* [26] add a constraint to avoid generating redundant overlapping proposals.

Since the advent of deep learning, the detection of action hypothesis has improved dramatically. We take inspiration from [14, 17], and use the detections generated by fine-tuning a CNN on the action categories present in a particular dataset.

**Association and labelling.** Some works [7, 14, 22] provide an evaluation on a per-frame basis. Without association, a robot would not be able to recognise that a walking action over multiple frames is in fact a single action[4]. Association in this context involves the assignment of detection-windows to existing action tubes. Most of the current state-of-the-art methods in action detection [7, 14, 17, 22] do link the detection-windows to generate space-time action tubes, however, so far this has been an offline procedure.

In the tracking community however, online data association methods are widely used. These include the multi-hypothesis tracker (MHT) [2, 15] and the Joint probabilistic data association (JPDA) algorithm [5, 8]. By design, most multi-target tracking methods [8, 15] are label-ignorant. They take detections with a single objectness-score as input, and associate them over time; previous action-detection works use these methods to associate detections of one class at a time. Our algorithm takes detections with a class-score vector (the direct output of most action/object-detection networks) as input. We tackle the problem of multi-label multi-target tracking, with the added complexity of dynamic labels for targets (shown in Fig.1 and 5). Thus we bridge the gap between human action detection [14, 17] and online multiple-object tracking [5], and propose a joint labelling and association algorithm inspired by [8].

Singh et al. [20] recently proposed an association algorithm for generating multiple action tubes incrementally. Note that [11, 18] also use modifications of this algorithm. Although their algorithm is online, it is not capable of handling multiple classes in a single pass, i.e they do the association independently for each action class. This causes the problem of multiple action tubes overlapping in the same space-time region even though only one action might be happening. This also makes their algorithm difficult to scale to large number of action categories. Also, they do a separate pass for temporal trimming. Whereas, we do the association, labelling and temporal localization in a single pass.

## 3 Methodology

Inspired by previous methods, we also first extract action detection-windows in each video frame independently (§ 4), and then perform linking of the detection-windows to construct action tubes, as detailed in what follows. A frame level action detection-window is a spatial bounding box enclosing a particular action category $l \in \mathcal{L}$, where $\mathcal{L}$ is the set of action categories; and an action tube represents a set of continuous detection-windows (without breaks) in time, which share the same action category $l \in \mathcal{L}$ [14, 17].

### 3.1 Online joint labelling and association (OJLA)

**Problem formulation.** At any time $t$, we seek to find the association between a set of currently active tubes $\{x_{t-1}^i\}$, $i \in \mathcal{N} = \{1, \ldots, N\}$, and a new set of detection-windows $\{y_t^j\}$, $j \in \mathcal{M}_0 = \{0, 1, \ldots, M\}$ from the current time-step, and also update the labels of the current tubes. Here '0' is a place holder for a dummy (or missed) detection. This is a one-to-one matching problem in which each tube should have a unique edge connecting it to a detection-window. The best assignment can be formulated as the solution to the problem of finding those connections which minimize an appropriate cost function $f$:

---

[4]Note that it is not easy for humans to determine whether an action detection system is connecting actions in time from a video displaying bounding box predictions, since we perform the data association effortlessly whilst watching.

$$\mathbf{a}^* = \arg\min_{\mathbf{a} \in \tau} f(\mathbf{a}), \tag{1}$$

where $\mathbf{a}$ is a binary assignment vector in which $a_i^j$ signifies that tube $i$ is connected to detection-window $j$, and $\tau$ is the space of all valid combinations of detection-to-tube assignments. Each valid assignment should satisfy the following constraints: i) each tube is uniquely assigned to a single detection (2a-i), and ii) each detection (except for dummy hypothesis $j = 0$) is assigned to at most one tube (2a-ii). Thus, $\tau$ is defined as the following set of binary vectors:

$$\tau = \Big\{ \mathbf{a} = (a_i^j)_{i \in \mathcal{N}, j \in \mathcal{M}_0} \,|\, a_i^j \in \{0, 1\}, \tag{2}$$

$$\text{such that (i)} \sum_{j \in \mathcal{M}_0} a_i^j = 1 \quad \forall i \in \mathcal{N}, \qquad \text{(ii)} \sum_{i \in \mathcal{N}} a_i^j \le 1 \quad \forall j \in \mathcal{M} \Big\}, \tag{2a}$$

where $a_i^j = 1$ indicates that tube $i$ is matched to detection $j$ and $a_i^j = 0$ indicates that tube $i$ is not matched to detection $j$. Note that $\mathcal{M}$ is the set of detection-window indices excluding the zero dummy index. Thus, $\mathbf{a} \in \tau \subseteq B^{N \times (M+1)}$ is a binary vector which represents one valid solution of the one-to-one matching problem.

**Cost formulation.** We denote the total cost for an assignment as $f(\mathbf{a}) = \mathbf{c}^\top \mathbf{a}$, where $\mathbf{c}$ is a cost vector in which $c_i^j$ is the cost for associating detection index $j \in \mathcal{M}_0$ with tube $i \in \mathcal{N}$ at time $t$. Finding the best assignment $\mathbf{a}^*$ thus becomes an integer linear program. For convenience, we define the cost $c_i^j$ in terms of a similarity score by inverting it:

$$c_i^j = \begin{cases} c_0, & \text{if } j = 0. \\ 1/s_i^j, & \text{otherwise,} \end{cases} \tag{3}$$

where $c_0$ is a constant for assigning the tube to no-detection, and $s_i^j$ is the score defining the similarity between tube $i$ and detection-window $j$.

A tube's state at time $t - 1$ is a composition of the bounding box coordinates, action scores, and it's action label:

$$x_{t-1}^i = (\mathbf{b}_{t-1}^i, \mathbf{z}_{t-1}^i, l^{i*}), \tag{4}$$

where $\mathbf{b}$ represents the detection-window parameters, $\mathbf{z}$ is a vector of action scores per category, and $l^*$ is the assigned action label. The detection-windows at time $t$ are composed only of boxes and scores: $y_t^j = (\mathbf{b}_t^j, \mathbf{z}_t^j)$.

The similarity score is defined in terms of a labelling score function $\psi$ and an overlap score function $\psi_o$:

$$s_i^j = \psi(x^i, y^j) + \psi_o(x^i, y^j), \tag{5}$$

where the overlap is calculated as the intersection-over-union between the tube and the detection-window.

The labelling score function is calculated as the maximum sum of unary tube and detection-window scores, minus a Potts penalty to encourage label smoothness, over all possible action labels:

$$\psi(x^i, y^j) = \max_{l \in \mathcal{L}} \Big( \mathbf{z}_{t-1}^i(l) + \mathbf{z}_t^j(l) - \bar{\psi}(l^{i*}, l) \Big). \tag{6}$$

Here $\mathbf{z}_{t-1}^i(l)$ is the unary score for the tube taking action label $l$, and $\mathbf{z}_t^j(l)$ is the corresponding score for the detection-window at time $t$. The label $l^{i*}$ is the assigned action label of the tube $i$, and $\bar{\psi}(l', l)$ is the Potts penalty incurred by switching from action $l'$ to $l$, where $\bar{\psi}(l', l) = \{0 \text{ if } l' = l, \quad 1 \text{ otherwise}\}$. The labelling term is motivated by the fact that we not only want to estimate the detection-window corresponding to the same tube, but also the action label which is being performed. Thus, Eqn. (6) decides the most probable action category assuming that detection $j$ forms part of tube $i$, taking care of the labelling whilst computing the score. Since the labelling score function makes action category decisions on each detection window, it will take care of the 'temporal trimming' of the action tube if a no-action category is present in the label set. In computing Eqn. (6), we found in practice that taking the sum of action scores from the tube over a time window of $n$ frames improves robustness to spurious predictions.

**Optimisation.** The optimal assignment $\mathbf{a}^*$ may be found using an off-the-shelf optimisation solver for integer linear programs. Instead of picking the best solution, [9] showed the improvements that can be obtained by marginalising over several (m-best) solutions.

## 3.2 Marginalizing over m-best

In [1, 9], the authors argue that in many cases it is beneficial to look at the m-best solutions of the one-to-one matching problem instead of just using the best solution. A brief motivation is that we are using a relatively simple cost function made up of unary and pairwise terms to model a complicated real-world application such as action detection. In practice picking the best solution may not always give good results, as there may be numerous competing solutions which are almost equally likely.

We therefore marginalize $f(\mathbf{a})$ over the m-best solutions of the matching space $\tau$. Let $\tau_i^j \subset \tau$ be the subset of $\tau$ which includes all the solutions in which tube $i$ is matched to detection $j$, the marginalized cost $q_i^j$ for assigning detection index $j \in \mathcal{M}_0$ to tube $i \in \mathcal{N}$ is calculated as:

$$q_i^j = -\log \sum_{\{\mathbf{a}_{\mathbf{k}}^* | \forall k \in [m], a_i^j = 1\}} e^{-f(\mathbf{a}_{\mathbf{k}}^*)}, \tag{7}$$

where $\mathbf{a}_{\mathbf{k}}^*$ is the $k^{\text{th}}$ best solution to Eqn. (1), as in [8]. The $k^{\text{th}}$ best solution is calculated using the binary tree partitioning (BTP) algorithm [8]. BTP removes the redundant constraints and computes the objective as a series of second-best solutions.

## 3.3 Tube-state update

The marginalised costs $(q_i^j)_{j \in \mathcal{M}_0}$ are normalised for each tube, and the detection corresponding to the minimum cost is assigned to tube $i$. Let $j^*$ be the index of the detection matched to tube $i$, and $l_t^*$ be the predicted action label calculated from Eqn. 6 for edge $(i, j^*)$. If the predicted action label is the same as the label of the action tube, the tube's state is then updated as: $x_t^i = (\mathbf{b}_t^{j^*}, \mathbf{z}_t^{j^*}, l^{i*})$. If $l_t^*$ is not the same as $l^{i*}$, the action tube is terminated, and a new tube with label $l_t^*$ is initiated.

**Initiation and termination.** In order to initiate and terminate action tubes, we took inspiration from a heuristic algorithm proposed for multi-target tracking [16]. Firstly, any detection-window that is not claimed by an existing action tube, but that has a high score for a particular action (excluding no-action) is initiated as a new tube. Secondly, a tube is terminated if its number of consecutive missed detection assignments reaches a specific threshold.

# 4  Frame Level Action Detections

For generating frame-level action detection-windows and scores we used the publicly available code of [17][5]. To demonstrate that our OJLA tube construction algorithm can be used to create a real-time action detection system, and to do a fair comparison of OJLA with the association algorithm of [20], we also pair it with the Single Shot Detector (SSD) [13] for frame-level action detections. Throughout the paper, we call this as SSD OJLA.

**Fusion of appearance and motion cues.**    Both the appearance and motion CNN networks output category-specific detection-windows and scores for each region proposal. We map the set of boxes and scores that originated from one region proposal back to a single box with a vector of scores by performing a weighted average of the boxes, where the weight corresponds to the action-specific score. The next step involves fusing detection-windows between the two independent appearance and motion CNNs, for which we use the union of the detections from appearance and motion streams.

# 5  Experiments and Results

**Improving computational performance.**    In order to improve computational performance, we make use of a gating procedure to eliminate unlikely detection-to-tube edges. In practice, we exclude the edges whose corresponding Euclidean distance between the tube and detection centers is very large ($> 10 pixels$). This translates into exploring only the local region around the tube's position in the previous frame. Although the detection network is trained with target action scores 0 and 1, all detection-windows above 0.9 are still initiated for higher recall. The threshold for terminating tubes is 5 consecutive missed detections, which takes care of spurious mis-detections by the network and short occlusions. The constant $c_0$ in Eqn. 3 is fixed to 10. Crucially, experiments confirm (details in supplementary material) that the algorithm is not sensitive to parameter changes. Also, these parameters are the same for all the 3 datasets namely J-HMDB-21, UCF-101 and LIRIS-HARL.

**Datasets.**    We evaluate our approach on three of the most challenging datasets on action detection, namely, UCF101 [21], J-HMDB-21 [10] and LIRIS-HARL [23]. In the J-HMDB-21 dataset each video only contains a single action which extends to the entire length of the video. We report results on J-HMDB-21 in Table 1 of the supplementary material. Providing more difficulty, the UCF-101 dataset contains sequences in which multiple action tubes of the same category are present simultaneously with varying temporal extent. Nevertheless, only one action category is present in each video. Our primary focus though is the LIRIS-HARL dataset, in which multiple action tubes, and action categories may be present simultaneously. In all of these three datasets, only one action is happening in a space time region. The qualitative results of our algorithm on videos in which different action categories occur concurrently are shown in Figs. 1, 5 and supplementary Fig.2

**OJLA with multiple labels for comparison.**    In our OJLA algorithm (§ 3), each tube is assigned only the label corresponding to the action with maximum score. With minor modification, our algorithm can also output several human-centered tubes, with each tube taking multiple labels. A tube is assigned label $l$ if its corresponding score in Eqn. 6 is above a threshold value. These two variants of OJLA are listed as 'Ours (OJLA)' and 'Ours (OJLA with multiple labels)' in Tables 1 and 2. OJLA with multiple labels is appropriate for less constrained scenarios where multiple actions co-occur in the same space-time region, for

*Table 1:* Quantitative action detection results UCF-101.

| mAP @ space-time overlap threshold $\delta$ | .2 | .5 | .6 | .75 | .5:.95 | Attributes |
|---|---|---|---|---|---|---|
| MR-TS R-CNN [14] | **73.50** | 32.10 | 11.25 | 02.70 | 07.30 | 1 – – 4 – – – – |
| Saha *et al.* [17] | 66.70 | 35.90 | 26.87 | 07.94 | 14.37 | 1 – – – 5 – – – |
| Ours (OJLA) | 59.58 | 31.13 | 24.70 | 06.29 | 12.70 | – 2 – – 5 – – – |
| Ours (OJLA with multiple labels) | 66.94 | **37.13** | **29.64** | **08.32** | **15.10** | – 2 – – 5 – – – |
| Yang *et al.* [24] | 73.67 | 37.80 | – | – | – | 1 – – – – – 7 – |
| Saha *et al.* (AMTnet) [18] | 63.06 | 33.06 | – | 00.52 | 10.72 | 1 – – – – – – 8 |
| Kalogeiton *et al.* [11] | **77.20** | **51.40** | – | **22.70** | **25.00** | – 2 – – – – – 8 |
| Singh *et al. et al.* [20] | **73.50** | **46.30** | – | **15.00** | **20.40** | – 2 – – – 6 – – |
| Ours (SSDapp. OJLA) | 55.41 | 31.09 | 26.13 | 12.14 | 14.59 | – 2 3 – – 6 – – |
| Ours (SSDapp. OJLA with multiple labels) | 68.32 | 40.52 | **34.06** | 14.27 | 18.55 | – 2 3 – – 6 – – |
| Ours (SSD OJLA) | 59.69 | 32.72 | 26.87 | 11.79 | 14.90 | – 2 – – – 6 – – |
| Ours (SSD OJLA with multiple labels) | 71.53 | 40.07 | 32.68 | 13.91 | 17.90 | – 2 – – – 6 – – |

(1) Offline, (2) Online, (3) Real-time,
(4) Faster R-CNN detections, (5) Fast-RCNN detections, (6) SSD detections, (7) CPLA, (8) multi-frame action tubelets

*Table 2:* Quantitative action detection results on LIRIS-HARL.

| mAP @ space-time overlap threshold $\delta$ | .4 | .5 | .75 | .5:.95 |
|---|---|---|---|---|
| Saha *et al.* [17] | 28.03 | 21.36 | – | – |
| Ours (OJLA) | **29.17** | **21.77** | **2.48** | **6.16** |
| Ours (OJLA with multiple labels) | 26.39 | 16.88 | 1.19 | 4.30 |

instance, when a person is walking and talking. It is noteworthy that our multilabel tube representation can differentiate between humans performing multiple actions and multiple humans performing individual actions, unlike previous methods [11, 14, 17, 18, 20].

**Results.** We compare the quantitative results that we achieved to the current state-of-the-art in Tables 1, 2. Despite using the same detection-windows and action scores as used in [17], and the fact that our method is incremental, our OJLA beats the state-of-the-art offline results [17] on the more difficult UCF-101 and LIRIS-HARL benchmarks. Moreover our algorithm is approximate given that we can only look at the present and past frames, whereas offline approaches can make use of the entire information carried by the test video.

In more experiments, we train the SSD network[6] [13] on UCF-101 using the same network architecture and training procedure as used in [13], and denote it as SSD OJLA. On UCF-101 our SSD OJLA performs at par with the state-of-art online association algorithm [20] in terms of accuracy, which also uses SSD detections; and outperforms it by huge margins in terms of association speed (550fps vs 400fps). Note however that [20] cannot handle multiple classes in a single pass, they perform association independently for each action class, and need a separate pass for temporal trimming. By contrast OJLA does the association, labelling and temporal localization in a single pass. Example qualitative results from UCF-101 are shown in Figs. 3 and 4. Note that our OJLA association algorithm can be used to get further gains in action detection performance by using more accurate tubelets [11] generated from a stack of frames, instead of frame-level detection windows. It is noteworthy that on LIRIS-HARL, a dataset which includes several action categories, our OJLA algorithm can truly leverage on its unique properties to get the best results across the board. Qualitative results from LIRIS-HARL are shown in Fig. 5 and in the supplementary material. We also report class-wise results on the LIRIS-HARL dataset in Table 3 of the supplementary mate-rial. Table 4 in the supplementary material shows the contribution of various components of our algorithm towards the results.

---
[6] https://github.com/weiliu89/caffe/tree/ssd

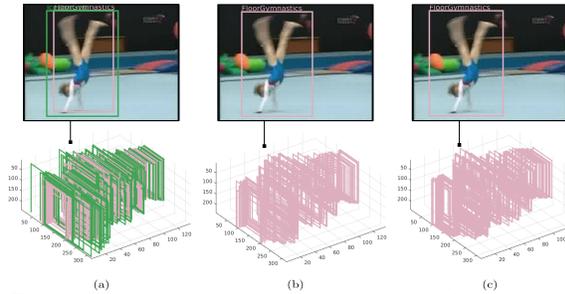

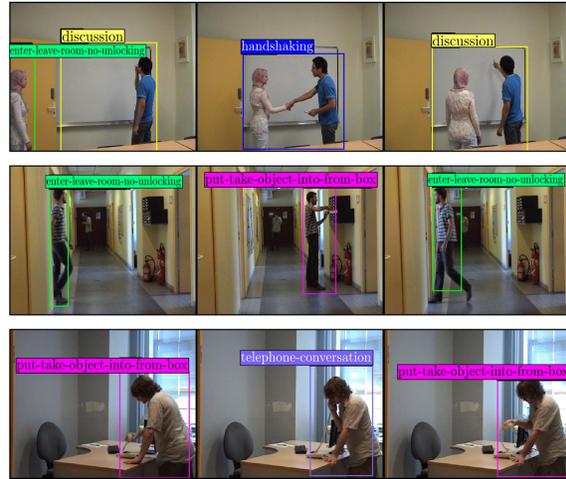

*Figure 3:* **(a)** Action tubes predicted by the method of [17] with 'ice dancing' (green) and 'gymnastics' (pink) labels in the same space-time region. **(b)** In contrast our method predicts tube with a single action label when only one action occurs in a space-time region, eliminating the possibility of predicting multiple co-located action tubes. **(c)** The ground truth action tube from UCF-101.

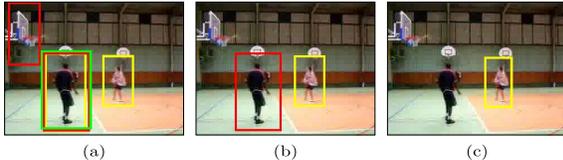

*Figure 4:* **(a)** Another example in which [17] predicts multiple co-located action tubes. This is a basketball sequence form UCF-101; the colours symbolise the following action categories: (green - Soccer Juggling), (red - Cricket Bowling), (yellow - Basketball). **(b)** Our predictions. **(c)** The ground truth annotation.

*Figure 5:* Qualitative results of our method on the LIRIS-HARL dataset. First (top), a woman walks into a room whilst a man stands in front of a whiteboard. The two people then 'shake hands' and start a 'discussion'. Notice how our algorithm is able to handle situations in which multiple actions occur concurrently and/or sequentially. Next (middle) a person 'enters/leaves a room without unlocking', then 'puts-takes an object from a box', and again 'enters/leaves a room without unlocking'. Finally (bottom) a man holds a 'telephone conversation'; again the system mislabels the beginning and end of the action by detection a 'put/take object into/from box' action immediately preceding and following the 'telephone conversation'.

**Discussion.** One of the main differences between our proposed method OJLA and those of previous methods [11, 17, 20] is that where appropriate (UCF-101, LIRIS-HARL), we can treat each region in space as belonging to one action category at one instant in time. In previous approaches, each spatial region assumes multiple action labels regardless. The difference this makes in terms of qualitative results can be seen in Figs. 3, 4 (also see supplementary Fig.1). For instance in Fig.3a, the method of [17] predicts that the girl performing a gymnastics action is also 'ice dancing', an impossible combination. Another example of the difference between these two approaches can be seen in Fig. 4, where two people are playing basketball. In this sequence, the person on the left is waiting to receive a ball and is not performing any action. However the multiple label approach of [17] labels the person with three actions; in many cases the fact that people are labelled with several actions can help to boost the results if at least one of the categories is correct (see Table 1). To prove the point, we implement a similar multiple label approach 'Ours (OJLA with multiple labels)' for comparison.

In JHMDB-21 and UCF-101 (scenarios where only one action is happening in a space-time region), even though one would expect the quantitative results of the multiple-label approach to become worse (see the qualitative comparison in Fig. 3) than 'OJLA', they in fact increase the mAP performance, as shown in Table 1. It is noteworthy that the two approaches (OJLA vs OJLA with multiple labels) being compared use the exact same association algorithm (OJLA). However, the multiple label approach predicts tube with multiple action labels whereas our OJLA algorithm predicts the exact same tube with just a single label. The reason for this discrepancy in results may be interpreted as follows. Consider a situation in which the detection-window association is accurate, but the labelling is poor. In this case a multiple label approach helps to improve results because predicting tube with multiple labels will increase the chances of getting a true positive detection. The single label approach, however, will suffer if the association is accurate but the labelling is poor, as its incorrect labelling (greedy) will not result in a true positive. We observed this behaviour on the J-HMDB-21 and UCF-101 datasets, where the mAP are quite high due to good detection-window scores and associations. When the association is less good, as observed on LIRIS-HARL, then predicting multiple action labels (as in the multilabel approach) won't help because neither of the tubes will have a correct label and all predictions would be counted as false positives (see Table 2). Thus, we can observe that predicting tubes with multiple labels can improve mAP results in relatively easy situations (indicating flaws with current action detection performance metrics), however, these improvements will not transfer to more complex scenarios.

**Real-time action detection.**   The SSD network takes ∼21.7ms per frame and OJLA takes ∼1.8ms per frame. The speed of the overall detection system is thus limited by the speed of the frame-level detection network, which is ∼46fps. The average linking speed in fps on UCF-101 of our method is 550fps, compared to 400fps of [20] and 300fps of [11] (supplementary Table 2). The enormous difference in speed is by virtue of our single pass formulation. [17, 20] perform separate passes for association and temporal localization, whereas our algorithm does both in a single pass.

# 6   Conclusion and Future work

In this work we proposed an algorithm called OJLA (Online Joint Labelling and Association) which improves over previous methods by doing away with multiple optimization passes for action tube window association, temporal localization and labelling. Instead, we formulated a novel cost function which solves these tasks jointly and incrementally. We demonstrated that OJLA has superior online association accuracy and speed as compared to the state-of-the-art action detection methods. In future work, we plan to explicitly model the space-time interplay between humans, actions and objects for a more detailed understanding of actions.

**Acknowledgements**   This work was supported by the EPSRC, ERC grant ERC-2012-AdG 321162-HELIOS, EPSRC grant Seebibyte EP/M013774/1, EPSRC/MURI grant EP/N019474/1 and the European Union's Horizon 2020 research and innovation programme under grant agreement No. 779813 (SARAS). Harkirat is wholly funded by a Tencent grant.